\documentclass{article}
\usepackage{kr}

\usepackage{times}
\usepackage{soul}
\usepackage{url}
\usepackage[hidelinks]{hyperref}
\usepackage[utf8]{inputenc}
\usepackage[small]{caption}
\usepackage{graphicx}
\usepackage{amsmath}
\usepackage{amsthm}
\usepackage{booktabs}
\usepackage{algorithm}
\usepackage{algorithmic}
\usepackage{researchpack}
\usepackage{xspace}
\usepackage{xcolor}
\usepackage{code}
\usepackage{hyperref}

\newcommand{\acronym}{NEuro-Symbolic consTrainEd structuRed predictor\xspace}
\newcommand{\method}{\textsc{Nester}\xspace}

\title{Neuro-Symbolic Constraint Programming for Structured Prediction}

\author{%
Paolo Dragone$^1$\and
Stefano Teso$^2$\and
Andrea Passerini$^2$ \\
\affiliations
$^1$Criteo\\
$^2$University of Trento\\
\emails
p.dragone@criteo.com,
\{stefano.teso, andrea.passerini\}@unitn.it
}

\begin{document}

\maketitle

\begin{abstract}
We propose \method, a method for injecting neural networks into constrained structured predictors. The job of the neural network(s) is to compute an initial, raw prediction that is compatible with the input data but does not necessarily satisfy the constraints. The structured predictor then builds a structure using a constraint solver that assembles and corrects the raw predictions in accordance with hard and soft constraints. In doing so, \method takes advantage of the features of its two components:  the neural network learns complex representations from low-level data while the constraint programming component reasons about the high-level properties of the prediction task.  The entire architecture can be trained in an end-to-end fashion. An empirical evaluation on handwritten equation recognition shows that \method achieves better performance than both the neural network and the constrained structured predictor on their own, especially when training examples are scarce, while scaling to more complex problems than other neuro-programming approaches. \method proves especially useful to reduce errors at the \emph{semantic} level of the problem, which is particularly challenging for neural network architectures.
\end{abstract}

\section{Introduction}
\label{sec:intro}

Neural networks have revolutionized several sub-fields of AI, including
computer vision and natural language processing.
Their success, however, is mainly limited to
``perception'' tasks, where raw unstructured data is abundant and
prior knowledge is scarce or not useful~\cite{darwiche2018human}.
Standard deep networks indeed rely
completely on inductive reasoning to acquire complex patterns from massive datasets.
This approach cannot be straightforwardly applied
to
machine learning problems characterized by
limited amounts of highly structured data.
Solving these tasks requires to generalize from few examples by leveraging
background knowledge, which in turn involves applying inductive \emph{and} deductive reasoning.
This observation has prompted researchers to develop neuro-symbolic models that integrate deep learning with higher level reasoning~\cite{de2020statistical,lake2017building,zambaldi2018relational,garnelo2016towards,santoro2017simple}.

We tackle neuro-symbolic integration in the context of structured output prediction under constraints.  In regular structured output prediction, the goal is to predict a structured object encoded by multiple interdependent and mutually constrained output variables~\cite{BakHofSchSmoTasVis07}.  Examples include parse trees, floor plans, game levels, molecules, and any other functional objects that must obey validity constraints~\cite{erculiani2018automating,di2020efficient}.

Our setting is more complex than regular structured prediction in that it involves predicting structures i)~subject to complex logical and numerical constraints, ii)~from sub-symbolic inputs like images or sensor data.
This problem is beyond the reach of existing methods.  Classical frameworks for structured prediction~\cite{lafferty2001conditional,tsochantaridis2004support} 
lack any functionality for representation learning, which is a key component for handling sub-symbolic data.
Most deep structured prediction approaches, on the other hand, do not support explicit constraints on the output, neither hard nor soft~\cite{belanger2016structured,amos2017input}.
Existing neuro-symbolic approaches combining logical constraints and deep learning either rely on fuzzy logic~\cite{diligenti2016learning,hu2016harnessing,donadello2017logic} or enforce satisfaction of constraints only in expectation~\cite{xu2017semantic}, in both cases failing to deal with hard constraints. Furthermore, these approaches are usually restricted to logical constraints.
The most expressive approach in this class is DeepProbLog~\cite{manhaeve2018deepproblog}, which enriches the probabilistic programming language ProbLog with neural predicates and is designed to process sub-symbolic data together with logic and (discrete) numeric constraints. However, probabilistic inference in DeepProbLog can become prohibitively expensive when dealing with structured prediction problems with a high degree of non-determinism, as highlighted by our experiments.

We propose \method (\acronym), a hybrid approach that integrates neural networks and constraint programming to effectively address constrained structured prediction with sub-symbolic inputs and complex constraints. \method combines a neural model with a constrained structured predictor into a two-stage approach.  On the one hand, the neural network processes low-level inputs (e.g., images) and produces a raw candidate structure based on the data alone.  This intermediate output acts as a initial guess and may violate one or more of the hard constraints.  On the other, the structured predictor refines the network's outputs and leverages a constraint satisfaction solver to impose hard and soft constraints on the final output.  Inspired by work on declarative structured output prediction~\cite{teso2017structured,dragone2018pyconstruct}, \method leverages the mid-level constraint programming language MiniZinc~\cite{nethercote2007minizinc} to define the constraint program, thus inheriting the latter's ability to deal with hard and soft constraints over categorical and numerical variables alike.  The entire architecture can be trained end-to-end by backpropagation.

\begin{figure*}[t]
\centering
        \includegraphics[width=0.95\textwidth]{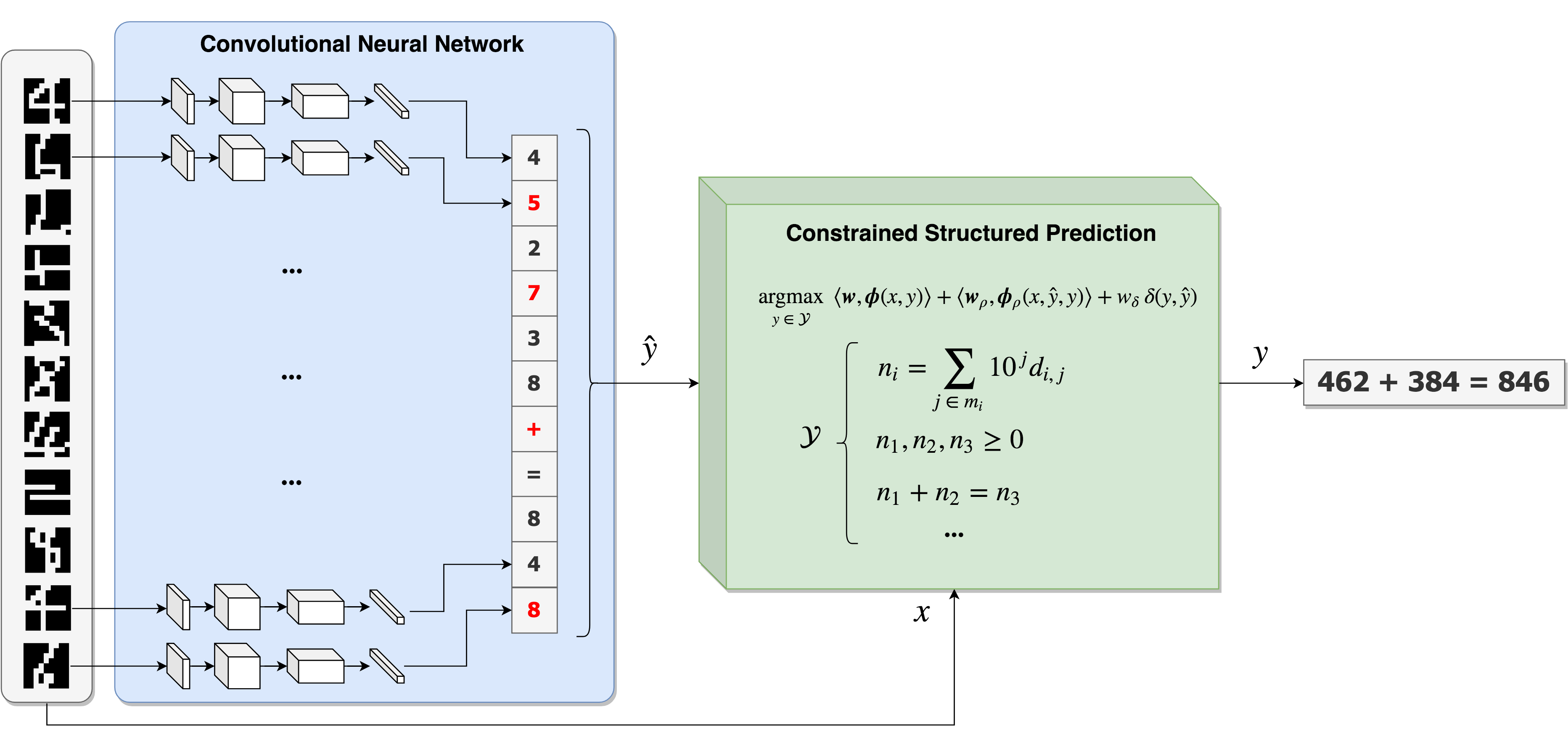}
\caption{\label{fig:schema} Architecture of the proposed model in the
  handwritten equation recognition setting. The figure shows the
  procedure for predicting an instance, i.e. a sequence of images
  forming an equation. Each image is first classified by a CNN (the
  same network for each image), and the predicted labels are grouped
  into a sequence $\hat{y}$. The structured predictor then takes the
  sequence $x$ and the intermediate prediction $\hat{y}$ and predicts
  a label $y$ by solving a structured prediction problem with the
  imposed constraints. Note how the network output $\hat{y}$ violates
  both syntactic and semantic constraints, while the final output $y$
  is guaranteed to satisfy both. Best viewed in colors.
}
\end{figure*}

We evaluated \method on handwritten \emph{equation} recognition~\cite{dragone2018pyconstruct}, a sequence prediction task in which the prediction must obey both syntactic and semantic constraints, cf. Figure~\ref{fig:schema}.
Our results show that the neural network and the constrained structured predictor work in synergy, achieving better results than either model taken in isolation.
Importantly, while the neural network alone can rather easily learn the correct {\em syntax} for the output, it struggles to achieve low error at the \emph{semantic} level of the problem.
Our combined architecture, on the other hand, outputs predictions that are both syntactically sound and semantically valid.

\section{Structured Prediction under Constraints}
\label{sec:pyconstr}

In structured prediction the goal is to predict an output
structure from an input, also typically structured. Standard examples
include part-of-speech tagging or protein secondary structure
prediction, where input and output are sequences of symbols, or parse
tree prediction, where a tree structure should be predicted starting
from an input sequence. Directly learning a mapping function
$f : \calX \to \calY$ is tricky given the structured nature of
$\calY$. Approaches to deal with the problem fall in two main
categories. Energy-based prediction models~\cite{LeCun06atutorial}
rely on a function $F : \calX \times \calY \to \bbR$ that computes the
compatibility between input and output. Learning aims at maximizing
the score (or minimizing the energy) of the correct output for a given
input, so that inference can be computed by maximizing the function
over the output space:
\begin{equation}
    \label{eq:structured_prediction}
    y = \argmax_{y' \in \calY} F(x,y')
\end{equation}
Search-based models~\cite{Dau2009,Ross2011} frame structured
prediction as a search in the space of candidate outputs, where a
scoring function evaluates the quality of a certain move (e.g.,
labeling the next output in the sequence) given the current state (the
input and the labels predicted for the previous
elements). Search-based models are typically more efficient than
energy-based ones, as they do not need to run inference over the entire
structure. On the other hand, by jointly inferring the entire
structure, energy-based models can naturally incorporate hard constraints
over the output structure. Indeed, while soft constraints between output
variables (e.g., the propensity for a verb to follow a noun) can be
implicitly learned by both framework in terms of weights associated to
states or input-output pairs, long-range hard constraints (e.g., non-overlap in furniture layouts~\cite{erculiani2018automating}, playability in game level generation~\cite{di2020efficient}) are challenging for
search-based models as they would require complex backtracking
operations
to retain satisfaction of the
constraint. The severity of the problem depends on the scope of the
constraint, with global (even soft) constraints being much harder to
manage than local ones. We thus focus on energy-based models in this
paper.

Structured SVM~\cite{tsochantaridis2004support,joachims2009predicting} is a popular
energy-based model\footnote{While energy-based models assume to
  minimize energy, structured SVM are typically defined in terms of
  scoring function (or negated energy) to be maximized. We stick to
  the standard formulation in this paper.}, that has the advantage of
carrying generalization guarantees of SVM to the structured prediction
setting. In structured SVM, $F(x, y)$ is modelled as a linear function of the type
$\inner{\vw}{\vphi(x, y)}$, where
$\vphi : \calX \times \calY \to \bbR^d$ is a feature map that
transforms input-output pairs into a $d$-dimensional joint feature
space, and $\vw\in\bbR^d$ is a parameter vector to be learned from
data. Given a training set of $n$ input-output pairs
$\{(x_i, y_i)\}_{i=1}^n$, where $y_i$ is a high-quality output for
$x_i$,
structured SVM learn $\vw$ by solving the following quadratic problem:
\begin{align}
\label{eq:ssvm}
\min_{\vw,\vxi} \quad & \frac{\lambda}{2}\norm{\vw}^2 + \frac{1}{n} \sum_{i=1}^n \xi_i \\
\text{s.t.} \quad     & \forall\ i\in[n], y\in\calY \setminus \{y_i\} \nonumber \\
                      & \inner{\vw}{\vphi(x_i, y_i)} - \inner{\vw}{\vphi(x_i, y)}\ge \Delta(y_i, y) - \xi_i \nonumber
\end{align}
The inequality constraints encourage the scoring function to rank the correct label above all the others.  In particular, for each training pair
$(x_i, y_i)$, they ensure that the score of the correct label
$\inner{\vw}{\vphi(x_i, y_i)}$ is larger than the score of any other
label by at least $\Delta(y_i, y)$, the latter being a task-dependent loss that
measures how much $y$ differs from $y_i$.  These constraints
are soft, and the slack variables $\vxi\in\bbR^n$ are penalties for
not satisfying them. The objective to be minimized combines the slacks
with and an $\ell_2$ regularization term
$\frac{\lambda}{2}\norm{\vw}^2$ that encourages simplicity and generalization.
Many algorithms have been developed for solving Eq.~\ref{eq:structured_prediction}
including
cutting plane~\cite{joachims2009cutting} and block-coordinate
Frank-Wolfe~\cite{lacoste2013block}. In this work we rely on
stochastic sub-gradient descent~\cite{ratliff2007approximate}, as it can be extended to deep models, as discussed in the next section. Stochastic
sub-gradient descent for structured SVM works by minizing the
\emph{structured hinge loss}:
\begin{equation*}
    L(\vw; x_i, y_i, y^*_i) = \frac{\lambda}{2}\norm{\vw}^2 + \inner{\vw}{\vphi(x_i, y^*_i) - \vphi(x_i, y_i)}
\end{equation*}
where
\begin{equation*}
    y^*_i = \argmax_{y\in\calY} \; \Delta(y_i, y) - \inner{\vw}{\vphi(x_i, y_i) - \vphi(x_i, y)}
\end{equation*}
Given $y^*_i$, it is possible to compute the sub-gradient
$\nabla_{\vw} L$ and then follow its direction via a standard gradient
descent step or variants thereof.

Depending on the model structure and output space $\calY$, the
hardness of computing a prediction (Equation~\ref{eq:structured_prediction})
varies greatly.
If the
output objects $y\in\calY$ are composed of real variables only and the
space $\calY$ is convex, then a prediction can be computed using gradient-based optimization methods.
Discrete variables render the task much more challenging. Most
research on structured prediction of discrete objects has focused on
problems for which an efficient prediction algorithm is known, such as
the Viterbi algorithm for sequences and the inside-outside algorithm
for trees, or has resorted to approximate inference~\cite{Fin2008} that does
not guarantee satisfaction of hard constraints. Following recent
improvements of constraint solving technologies, it has become
increasingly more feasible to delegate the job of solving the
inference problem in Equation~\ref{eq:structured_prediction} to a
generic constraint solver~\cite{teso2017structured}. Nowadays, mixed
integer linear programming frameworks can solve optimization problems
with several hundred variables in a matter of milliseconds.
Constraint programming languages such as
MiniZinc~\cite{nethercote2007minizinc} have also been leveraged in
structured prediction problems involving complex
output structures~\cite{dragone2018pyconstruct}, such as layout
synthesis~\cite{erculiani2018automating} and product bundling
recommendation~\cite{dragone2018no}. Constraint
programming is particularly well suited
for encoding structured prediction tasks, as it
allows to easily embed background knowledge directly into the
predictor in the form of arbitrary logic and algebraic constraints. For instance,
in the equation recognition task described in this paper, we are
able to easily encode the fact that equations must be valid by specifying constraints
that transform the sequence of labels into integers and another constraint defining
the relationship between these integers (more details in Section~\ref{sec:experiments}).

\section{Neural Constrained Structured Prediction}
\label{sec:neural_structured}

A substantial limitation of approaches based on structured SVM is that
they are shallow, leaving to the user the burden of designing
appropriate joint feature mappings $\phi$. On the other hand, the
ability to learn effective representations from low-level data is a
key reason for the success of deep learning technology. Indeed, deep
structured prediction has become predominant in domains like
NLP~\cite{OtterMK21} which are characterized by an abundance of
labelled data. The standard solution to apply neural approaches to
structured prediction problems is that of search-based learning, where
auto-regressive models like LSTM or attention-based models are
employed to predict the next output given the subset of already
predicted ones. This solution however suffers from a number of
problems~\cite{deshwal2019learning}, like the so-called {\em exposure bias} (during
training, the ground-truth is assumed for the subset of already
predicted variables) and the discrepancy between the training loss
(cross entropy on the label of the next output) and the task loss
(quality of the overall output prediction). While several ad-hoc
solutions have been developed to try addressing these problems, by
adapting techniques developed for shallow search-based structured
prediction~\cite{Ross2011}, these approaches share with their shallow
counterparts the lack of support for hard constraints. Deep
energy-based approaches have also been
explored~\cite{belanger2016structured,gygli2017deep}. However they rely on
gradient-based inference in a relaxed output space, and thus again
they cannot satisfy constraints. Indeed, satisfaction of constraints
has been indicated as one of the major challenges for deep structured
prediction~\cite{deshwal2019learning}.

To solve this issue, we propose \method, a neural constrained
structured prediction model. \method features a general-purpose
constrained structured predictor~\cite{teso2017structured} (see
Section~\ref{sec:pyconstr}) that acts as a ``refinement'' layer on top
of the predictions made by a standard neural network. The constrained
structured predictor is fed both the input $x\in\calX$ and the
predictions of the neural network $\hat{y}\in\calY$, and outputs a
final prediction $y\in\calY$.  Figure~\ref{fig:schema} shows, as an
example, the architecture that has been used in our experiments on
handwritten equation recognition (see Section~\ref{sec:experiments}),
but different architectures may be used depending on the domain. Note
that the neural network does not have to be aware of the structure of
the output, and may even simply predict the output variables
independently.  The constrained structured predictor is in charge of
grouping the predictions into the desired structure while
``correcting'' the mistakes of the network and enforcing the
constraints of the problem. To do so, both hard and soft constraints
are encoded in the constraint program. The former represent hard
requirements that the output needs to satisfy in order to be
considered valid (e.g., algebraic constraints for handwritten equation
recognition). The latter represent features defined over a combination
of inputs, network outputs and refined outputs, which potentially
correlate with the correctness of the refinement, and whose weights
are learned during training.

More specifically, prediction consists in solving the following
constrained optimization program:
\begin{eqnarray*}
  \label{eq:argmax}
    \argmax_{y \in \calY} & & \inner{\vw}{\vphi(x, y)} + \inner{\vw_{\rho}}{\vphi_{\rho}(x, \hat{y}, y)} + w_{\delta} \, \delta(y, \hat{y})  \\
  \textrm{s.t.} & & \hat{y} = g(x;W) 
\end{eqnarray*}
where $\calY$ encodes the hard constraints, $g(x;W)$ is an arbitrary
neural network (ignoring the constraints), $\vphi(x, y)$,
$\vphi_{\rho}(x, \hat{y}, y)$ and $\delta(y, \hat{y})$ are features,
and $(\vw,\vw_{\rho},w_{\delta})$ and $W$ are learnable weights of the
constrained structured predictor and the neural network
respectively. We refer to $\vphi(x, y)$ as \emph{prediction features},
whereas $\vphi_{\rho}(x, \hat{y}, y)$ are named \emph{refinement
  features}. The former are features that help predicting the right
output from the input, analogously to the ones used in standard
structured prediction models~\cite{lafferty2001conditional}. The
latter are features that should help spotting the common mistakes made
by the network, by relating input, output and network
predictions. Finally $\delta(y, \hat{y})$ represents a distance
measure between the prediction of the structured predictor and the one
of the neural network and, depending on $w_{\delta}$, acts as a
regularizer, preventing the structured predictor from learning to make
predictions too dissimilar from the neural network.

Learning the combined model can be done in three steps:
\begin{enumerate}
    \item If possible, pre-train the neural network; as mentioned, the neural network does not have to be aware of the output structure, so if the structured labels can be deconstructed into the basic output components of the neural network then the network can be pre-trained using those.
    \item Train the constrained structured prediction model. At this stage, the structured predictor is trained independently from the neural network and it gets the predictions of the neural network as if they were any other inputs.
    \item Fine tune the constrained structured predictor and the neural network end-to-end, using stochastic sub-gradient descent and back-propagating the gradient of the structured hinge loss through the structured model and through the network.
\end{enumerate}

The last point can be achieved by simply chaining the gradients of the structured hinge loss with respect to $\hat{y}$ and the gradient of the neural network output with respect to its weights. For the last layer of the network with weights $W_K$:
\begin{equation*}
    \nabla L = \left[\frac{\partial L}{\partial \hat{y}}\frac{\partial \hat{y}}{\partial W_{K}}\right]
\end{equation*}
Gradients for the other layers can be obtained by standard backpropagation through the network, chaining gradients over the layers starting from the last one.

\begin{figure*}[t]

\centering
\begin{tabular}{cc}
\hspace{-0.1in}
\includegraphics[width=0.5\textwidth]{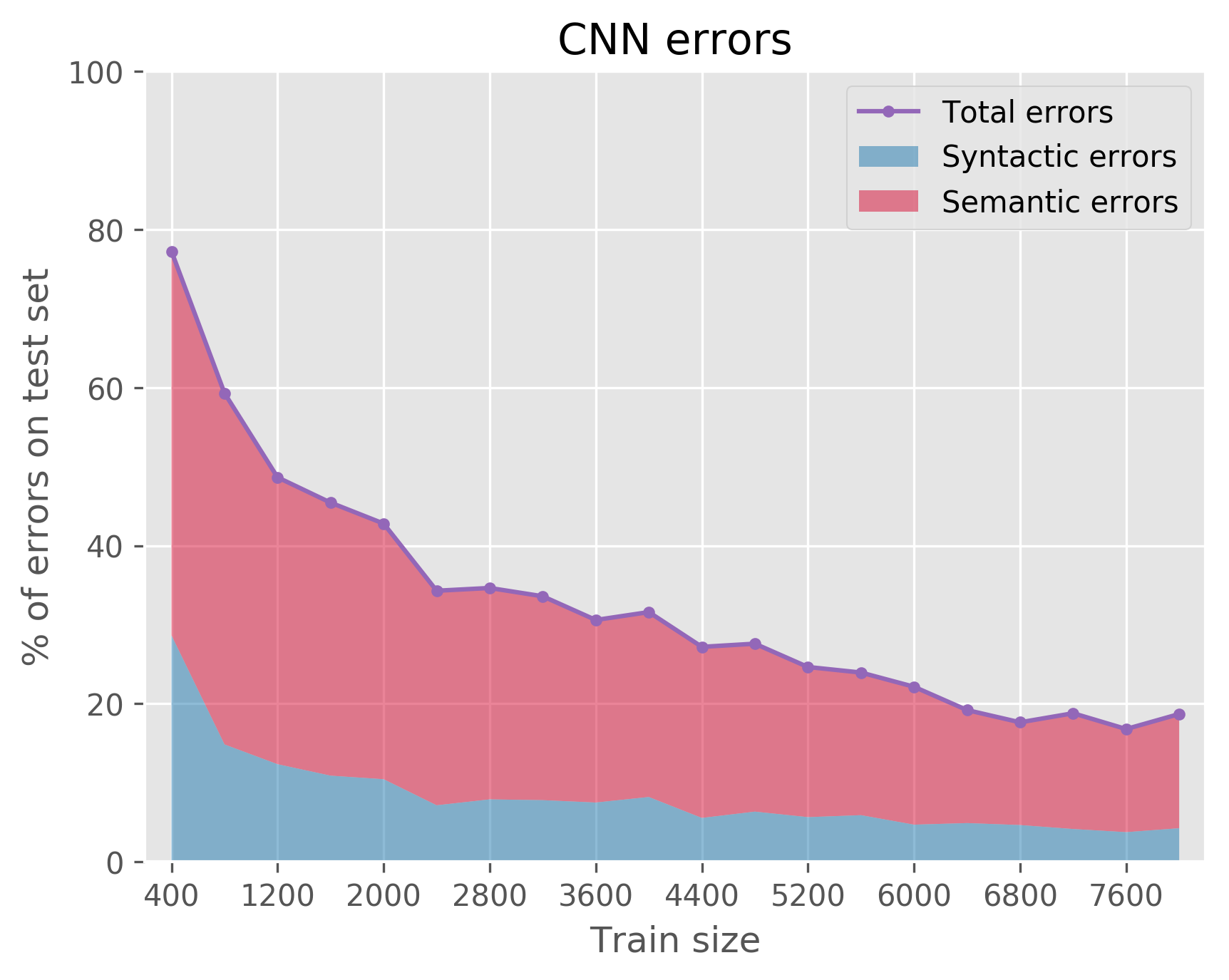}
&
\hspace{-0.2in}
\includegraphics[width=0.5\textwidth]{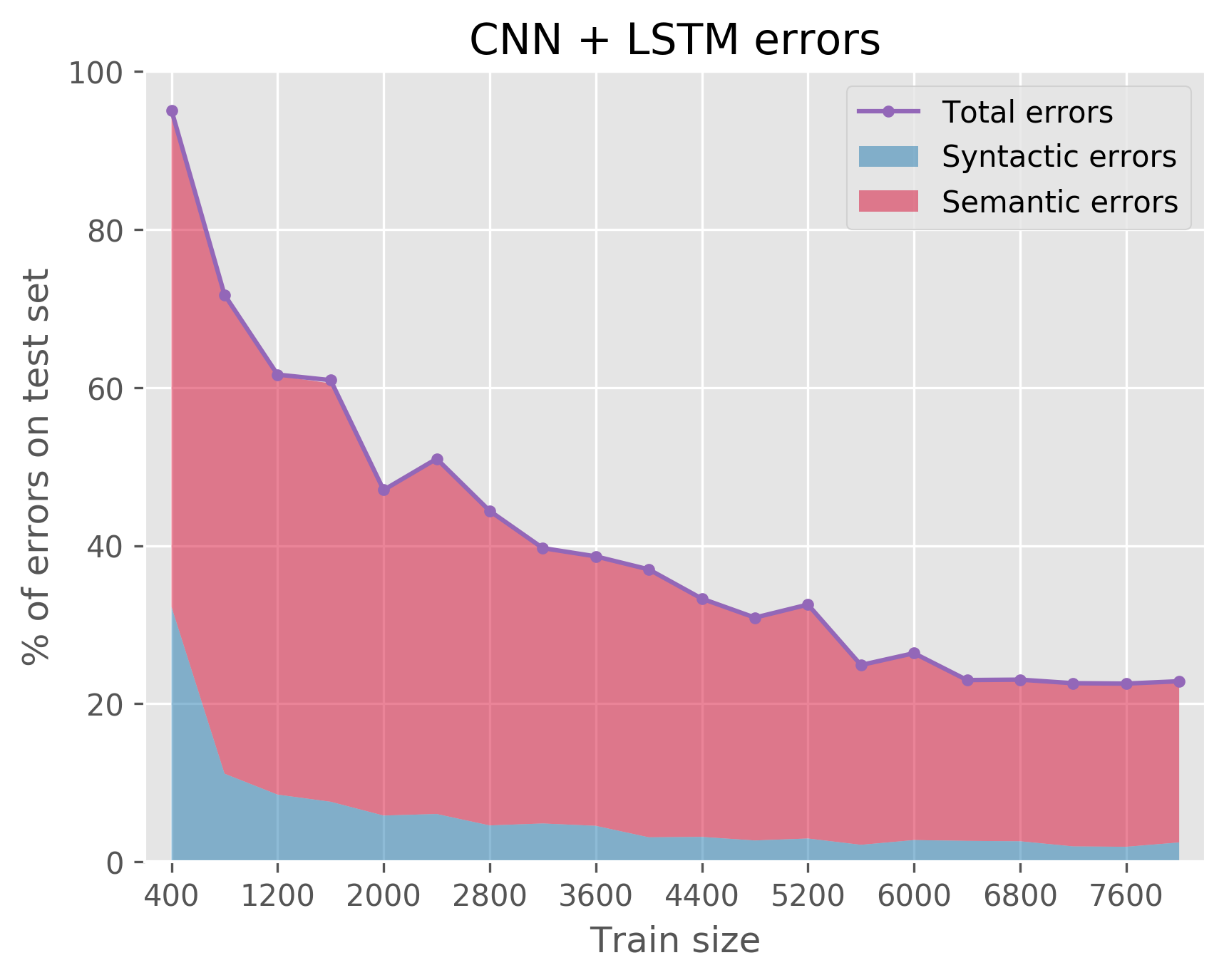}
\end{tabular}

\caption{
\label{fig:percent_errors}
Percentage of erroneously predicted equations by the CNN (left) and by the CNN +
LSTM (right) on the test set with increasing training set size. The plots show
the percentage of total errors over the test set size and its decomposition into
syntactic and semantic errors.  Best viewed in colors.
}
\end{figure*}

\section{Neural constrained sequence prediction}

In the rest of the paper, we will focus on a common kind of structured
prediction problem, namely sequence prediction.
In (discrete) sequence prediction, the objects to predict are sequences of
varying length $m$, i.e. $y=(y_1,\dots,y_m)$, and each element $y_j$ of the
sequence takes values from an alphabet of $q$ possible symbols
$\{s_k\}_{k=1}^q$.  For each element of the sequence, a vector of $l$ input
features is typically available. We indicate with $x_{e,i}$ the $i$-th feature
of the $e$-th element of the input sequence.

As prediction features we employ standard features for correlating input and
output variables commonly used in e.g.  conditional random
fields~\cite{lafferty2001conditional} for sequence prediction.  These come in
the form of two vectors, and are referred as \emph{emission} and
\emph{transition} features. The former are encoded as:
\begin{equation}
\phi_{i, k}(x, y) = \sum_{e = 1}^m x_{e, i} \cdot \llbracket y_e = s_k \rrbracket
\end{equation}
where $i$ ranges over input features and $k$ over output symbols, and the
expression $\llbracket \cdot \rrbracket$ equals to $1$ if the argument is true,
$0$ otherwise. The resulting vector can then be used to correlate the appearance
of a specific pixel inside input images with each of the emitted symbols.  The
transition features are define as follows:
\begin{equation}
 \phi_{k_1, k_2}(x, y) =
    \sum_{e = 1}^{m - 1} \llbracket y_e = s_{k_1}\ \land\ y_{e + 1} = s_{k_2} \rrbracket
\end{equation}
where $k_1, k_2$ ranges over all the possible combinations of couples of
symbols, counting the number of times in which two symbols appear one after the
other. This formulation enables the predictor to correlate the appearance of
consecutive symbols inside the output sequence.

The refinement features, on the other hand, need to correlate mistakes of the
network with the input and output variables. For this sequence prediction task,
we define the refinement features vector $\vphi_{\rho}$ as:
\begin{equation}
    \label{eq:sequence_features}
    \phi_{i, k}(x, \hat{y}, y) = \sum_{e = 1}^{m} x_{e, i} \cdot \llbracket y_e = s_k \ \land \ \hat{y}_e \neq s_k \rrbracket
\end{equation}
where $i$ ranges over input features and $k$ over output symbols.  These
features basically correlate the values of each input feature with the
overriding of neural network predictions by the refinement layer. By learning
appropriate weights for them, the structured predictor can learn to fix the most
common mistakes made by the neural network.
Finally, we define $\delta(y, \hat{y})$ as the Hamming distance between $y$ and $\hat{y}$:
\begin{equation*}
    \delta(y, \hat{y}) = \sum_{e=1}^m \llbracket y_e \neq \hat{y}_e \rrbracket
\end{equation*}
Table~\ref{tab:features_constraints} contains an overview of all the components
of the constrained problem we used in our experiments.

\section{Experiments}
\label{sec:experiments}

We tested our technique on the handwritten equation recognition task described
in~\cite{dragone2018pyconstruct}. All our experiments are implemented
using
Tensorflow and Pyconstruct~\cite{dragone2018pyconstruct}.
MiniZinc~\cite{nethercote2007minizinc} was used as constraint programming engine
and Gurobi\footnote{\url{http://gurobi.com}.} as the underlying
constraint solver.
The code of \method will be made available upon acceptance.

\subsection{Setting}

Each input object is a sequence of images of variable length, and the
corresponding output is a sequence of symbols of the same length. The
possible symbols include the digits from $0$ to $9$, $+$ and $=$. The
equations are all of the form $a + b = c$, where $a, b, c$ are
arbitrary positive integers (of a predefined maximum number of
digits), and the equations are all \emph{valid}, meaning that it is
always true that $a$ plus $b$ equals $c$. This is the background
knowledge of the problem. Images are matrices of $9 \times 9$ black
and white pixels. The dataset was constructed by assembling 10,000
valid equations from the ICFHR'14 CROHME competition data.
All our experiments were run on the same train-test split,
consisting of 8,000 equations for training and 2,000 for
testing. To highlight the behavior of the models with
different amounts of training examples, we divided the training set
into 20 chunks of increasing size and report results over the test set
for each training set chunk.

\subsection{Neural model}
As highlighted in Figure~\ref{fig:schema}, we use a CNN to predict the symbol of
each image in the sequence independently. The CNN is composed of two
convolutional layers with $3 \times 3$ filters and ReLU activation, each
followed by a $2 \times 2$ max-pool layers, then a $128$ dense layer with
dropout regularization ($0.5$ probability), and finally a softmax layer with
$12$ outputs, one for each symbol. The CNN is trained using Adam~\cite{adam}
with a cross entropy loss.

The left plot in Figure~\ref{fig:percent_errors} reports the relative
misclassification error of the CNN over the test set, i.e. the percentage of
sequences for which the network made at least one mistake. The violet line
indicates the total percentage of errors across the various training set chunks.
As the network does not enforce any constraint on the output, it is subject to
errors resulting from predicting inconsistent output objects. We subdivide the
constraint satisfaction errors into \emph{syntactic} errors, for which the
prediction is not properly formatted according to the template $a + b = c$, and
\emph{semantic} errors, i.e. well-formatted predictions for which the resulting
equation is not valid, that is $a$ plus $b$ does not equal $c$. Occasionally the
network makes mistakes that are not due to unsatisfied constraints, but these
are a very slim minority. The shaded areas in Figure~\ref{fig:percent_errors}
show how the different types of errors contribute to the total. While the total
number of errors decreases with increasing training set size, the proportion
between syntactic and semantic errors settles around $33\%$ to $67\%$,
indicating that semantic errors are generally harder to correct.
To make sure that these errors are not simply due to the fact that the
CNN predicts each output symbol independently, we also experimented with a
recurrent architecture. The recurrent network is assembled by stacking over each
convolutional layer of our CNN an LSTM layer.  Differently from the CNN, the
network is trained with whole input-output sequence pairs.

The right plot in Figure~\ref{fig:percent_errors} shows the errors of
this architecture and their decomposition into syntactic and semantic
types.  Overall, the network made more errors, possibly due to the
increased number of trainable parameters and thus the necessity for
more examples to be properly trained. Relatively speaking, though, the
CNN + LSTM network learned very quickly how to fix syntactic errors,
even faster than the CNN. Yet, semantic errors were much more
difficult to correct, and they ended up being the large majority of
the errors of the network. These results show that even highly-expressive
neural network architectures can have problems in going beyond the
shallow syntactic level and reach a deeper semantic level of
understanding, especially when limited data is available.

\subsection{Constrained structured model}
\label{par:eval_cst}
As constrained structured layer (``CST'' from now on) we used an
enhanced version of the constrained model
from~\cite{dragone2018pyconstruct}. Not being conceived to refine
predictions from other modules, the original model only contains
prediction features $\vphi(x, y)$.  Our model extends the original one
by taking the prediction of the neural network $\hat{y}$ as input,
combining the prediction features with the refinement features
described in Equation~\ref{eq:sequence_features}, and including the
Hamming distance between $y$ and $\hat{y}$ (see
Figure~\ref{fig:schema}).
A summary of the features and constraints
used in the structured predictor is given in
Table~\ref{tab:features_constraints}.

\begin{table}
\caption{
\label{tab:features_constraints}
Summary of the components of the constrained structured predictor in the
handwritten equation recognition setting. The expression $\llbracket \cdot
\rrbracket$ equals to $1$ if the argument is true, $0$ otherwise.
}
\centering
\begin{small}
\def\arraystretch{1.38}
\begin{tabular}{l}
    \hline
    \textbf{Input} \ $x, \hat{y}$\\
    $\circ\ $ Array of $m$ \ $9\times9$ images, of $\{0, 1\}$ pixels \vspace{0.02in} \\
    \qquad $x \in  \{0,1\}^{m\times9\times9} $ \\
    \quad where \\
    \qquad $x_{e, i, j} \in \{0, 1\}$ \quad $(i, j)$-th pixel of the $e$-th image \\
    $\circ\ $ Array of $m$  predictions of the neural network (NN)  \vspace{0.02in} \\
    \qquad $\hat{y} \in  \{0,\dots,q\}^{m} $ \\
    \quad where \\
    \qquad $q = 11$ $\implies$ digits: 0-9, \ plus ($+$): $10$, \ equals ($=$): $11$ \\
    \qquad $\hat{y}_{e} \in \{0,...,q\}$ \quad NN prediction for the $e$-th image \\[3pt]
    \hline
    \textbf{Output} \ $y$ \\
    $\circ\ $ Array of $m$ output symbols \vspace{0.02in} \\
    $\qquad y \in  \{0,...,q\}^{m} $ \\
    \quad where \\
    \qquad $q = 11$ $\implies$ (same as above) \\
    \qquad $y_{e} \in \{0,...,q\}$ \quad output symbol for the $e$-th image \\[3pt]
    \hline
    \textbf{Prediction features} \ $\vphi(x, y)$\\
    $\circ\ $ \emph{Emission features: } appearance of a pixel in images of a class, \\
    $\qquad \phi_{i, j, k}(x, y) = \sum_{e = 1}^m x_{e, i, j} \cdot \llbracket y_e = k \rrbracket$ \vspace{0.02in} \\
    \quad where \\
    \qquad $k \in \{0, \dots, 11\}$ ranges over symbols \\
    \qquad $(i, j) \in \{0, \dots, 9\}^2$ range over pixels \\
    $\circ\ $ \emph{Transition features: } classes in contiguous images, \\
    $\qquad \phi_{k_1, k_2}(x, y) = \sum_{e = 1}^{m - 1} \llbracket y_e = k_1\ \land\ y_{e + 1} = k_2 \rrbracket$ \vspace{0.02in} \\
    \quad where \\
    \qquad $(k_1, k_2) \in \{0, \dots, 11\}^2$ range over pairs of symbols \\[3pt]
    \hline
    \textbf{Refinement features} \ $\vphi_\rho(x, \hat{y}, y)$ \\
    $\circ\ $ Disagreement between the network prediction and the output, \\
    $\qquad \phi_{i, j, k}(x, \hat{y}, y) = \sum_{e = 1}^{m} x_{e, i, j} \cdot \llbracket y_e = k \ \land \ \hat{y}_e \neq k \rrbracket$ \vspace{0.02in} \\
    \quad where \\
    \qquad $k \in \{0, \dots, 11\}$ ranges over symbols \\
    \qquad $(i, j) \in \{0, \dots, 9\}^2$ range over pixels \\[3pt]
    \hline 
    \textbf{Distance measure} \ $\delta( \hat{y}, y)$\\
    $\circ\ $ Hamming distance between output and network prediction \\
    $\qquad  \delta(y, \hat{y}) = \sum_{e = 1}^m \llbracket y_e \neq \hat{y}_e \rrbracket$ \vspace{0.02in} \\[3pt]
    \hline
\end{tabular}
\end{small}
\end{table}

The definition of the constrained structured model is presented in the MiniZinc
code listed in Figure~\ref{fig:code}.  The sequence of images composing the
equation is given as input. The output consists of a sequence of symbols equal
to the length of the input sequence, each symbol being an integer between zero
and eleven (0-9 for the digits, 10 and 11 for the $+$ and $=$ operators
respectively). The input also contains the sequence of symbols predicted by the
CNN.

\begin{figure}
\begin{lstlisting}[language=PyMzn]
%% ~~~~~~~~~~~~~~~~~~~~~~~~~~ INPUT/OUTPUT ~~~~~~~~~~~~~~~~~~~~~~~~ %%
%% Max digits per number, dimension of the images (9 x 9)
int: DIGITS = 3;
set of int: HEIGHT = 1 .. 9; set of int: WIDTH = 1 .. 9;

%% Assume "+" is encoded by 10 and "=" is encoded by 11
set of int: SYMBOLS = 0 .. 11; int: PLUS = 10; int: EQ = 11;

%% INPUT: sequence length, tensor of images, CNN prediction
int: len; set of int: SEQ = 1 .. len;
array[SEQ, HEIGHT, WIDTH] of {0, 1}: img;
array[SEQ] of SYMBOLS: cnn_seq;

%% OUTPUT: sequence of symbols
array[SEQ] of var SYMBOLS: seq;

%% ~~~~~~~~~~~~~~~~~~~~~~~~ DOMAIN KNOWLEDGE~~~~~~~~~~~~~~~~~~~~~~~ %%
%% Indices of the two operators
array[1 .. 2] of var 2 .. (length - 1): opr;

%% Syntactic constraints: exactly one "+" and one "=", "+" before "=",
%% positions corresponding to indices "opr" in the sequence
constraint count(seq, PLUS, 1) /\ count(seq, EQ, 1);
constraint increasing(opr) /\ seq[opr[1]] == PLUS /\ seq[opr[2]] == EQ;

%% Extremes and number of digits of the three numbers
array[1 .. 4] of var 0 .. len + 1: ext = [0, opr[1], opr[2], len + 1];
array[1 .. 3] of var 1 .. DIGITS: n_digits = [
    ext[i + 1] - ext[i] - 1 | i in 1 .. 3 ];

%% Matrix of digits, zero-padded on the left if less than three digits
array[1 .. 3, 1 .. DIGITS] of var SYMBOLS: digits =
    array2d(1 .. 3, DIGITS, [
        if j <= (DIGITS - n_digits[i]) then 0 else
            seq[ext[i] + j + n_digits[i] - DIGITS]
        endif
    | i in 1 .. 3, j in 1 .. DIGITS ]);

%% The three numbers, sum digits for corresponding power of 10
array[1 .. 3] of var int: num = [
        sum(j in 1 .. DIGITS)(pow(10, DIGITS - j) * digits[i, j])
    | i in 1 .. 3 ];

%% Semantic constraint: the equation must be valid
constraint num[1] + num[2] == num[3];

%% ~~~~~~~~~~~~~~~~~~~~~~~~~~~~ FEATURES ~~~~~~~~~~~~~~~~~~~~~~~~~~~~ %%
%% EMISSION + TRANSITION + REFINEMENT + DISTANCE
int: N_FEATURES = 9*9*12 + 12*12 + 9*9*12 + 1;

array[1 .. N_FEATURES] of var int: features =
   %% EMISSION FEATURES
   [ sum(e in SEQ)(img[e, i, j] * (seq[e] == s))
    | i in HEIGHT, j in WIDTH, s in SYMBOLS ]
   %% TRANSITION FEATURES
++ [ sum(e in 1..len - 1)(seq[e] == s1 /\ seq[e + 1] == s2)
    | s1, s2 in SYMBOLS ]
   %% REFINEMENT FEATURES
++ [ sum(i in SEQ)(img[i, j, k] * (seq[i] == s) * (cnn_seq[i] != s))
    | j in HEIGHT, k in WIDTH, s in SYMBOLS ]
   %% HAMMING DISTANCE
++ [ -sum(i in SEQ)(cnn_seq[i] != seq[i]) ];
\end{lstlisting}
    \caption{\label{fig:code} A sketch of the MiniZinc program for the
    constrained structured model for the equation recognition problem. The
    MiniZinc program encodes the input images tensor, and neural network
    predictions (lines 10-12) and the output sequence of symbols (line 15), the
    domain knowledge in the form of hard constraints ensuring the equation is
    properly syntactically formatted (lines 18-29) and enforcing the semantic
    validity of the equation (lines 31-45), and finally the features of the
    structured model as defined in Equation~\ref{eq:argmax}, prediction features
    (lines 52-57), refinement features (lines 58-60) and distance metric (lines
    61-62), whose weights are learn from data. For brevity, the figure omits
    some parts regarding the definition of the weights and the objective
    function. Best viewed in color.}
\end{figure}

Syntactic constraints over the output are encoded in lines $23$ and $24$. They
enforce the presence of a single $=$ and a single $+$ in the output sequence,
plus the constraint that the $+$ symbol should come before the $=$ one. The next
code section (from line $26$ to $42$) is devoted to converting the sequence of
output symbols into an actual algebraic formula. This procedure starts in line
$27$ with the definition of the extremes of the three numbers, that are given by
the two operators. The extremes are used to count the number of digits that
compose each of the three numbers (lines $28$-$29$). These two pieces of
information, coupled with the sequence predictions, are then used in lines
$32$-$37$ to populate a $3\times3$ matrix, where each row represents one of the
three numbers, and the columns represent the digits, sorted from the most
significant to the least significant. Each number can be composed at most by $3$
digits, if the digits are less than $3$, the number is left padded with the
appropriate number of zeros. The matrix is then transformed into $3$ integer
values $a$, $b$ and $c$ (lines $40$-$42$), summing up all digits of each row,
multiplied by the appropriate power of $10$, according to their relative
position. In formula, each row $i$ is converted into an integer $n_i = \sum_{0
\le j < m_i} 10^j \cdot d_{i, j}$, where $m_i$ is the number of digits in
number $i$ and $d_{i,j}$ is the $j$-th digit of the $i$-th number, from the
least to the most significant digit. This conversion allows us to impose in line
$45$ the semantic constraint enforcing the output sequence to encode a valid
formula, i.e. $a + b = c$.

The last section of the code is devoted to declaring the features, with the
formulation introduced in Section~\ref{sec:neural_structured} (from line $47$ to
$62$). This code is a template that Pyconstruct~\cite{dragone2018pyconstruct}
uses to solve different inference problems during training and prediction. At
runtime, the library takes care of determining which objective function to
optimize depending on the inference problem at hand.

\begin{figure*}[t]

\centering
\begin{tabular}{cc}
\hspace{-0.1in}
\includegraphics[width=0.5\textwidth]{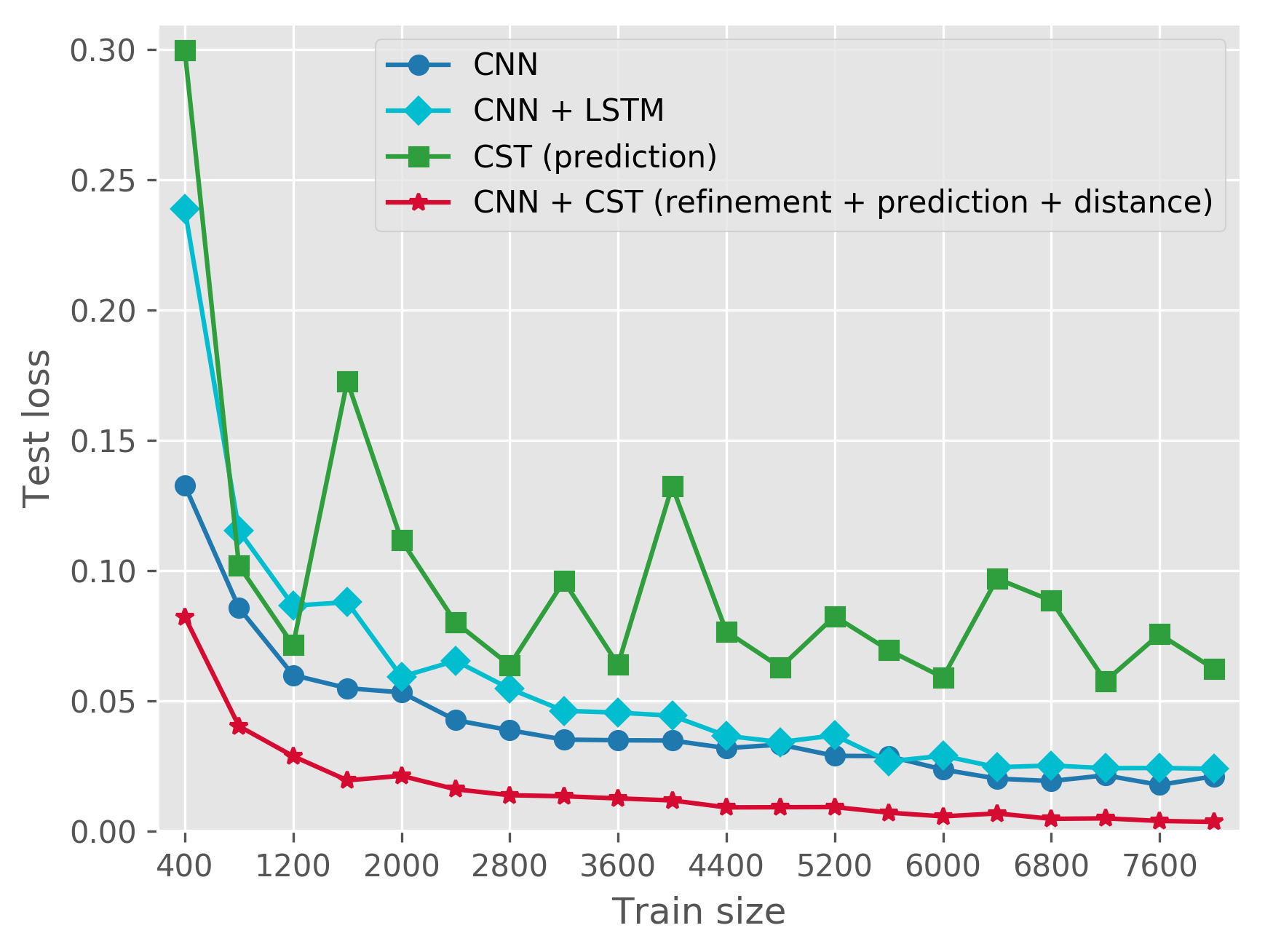}
&
\hspace{-0.2in}
\includegraphics[width=0.5\textwidth]{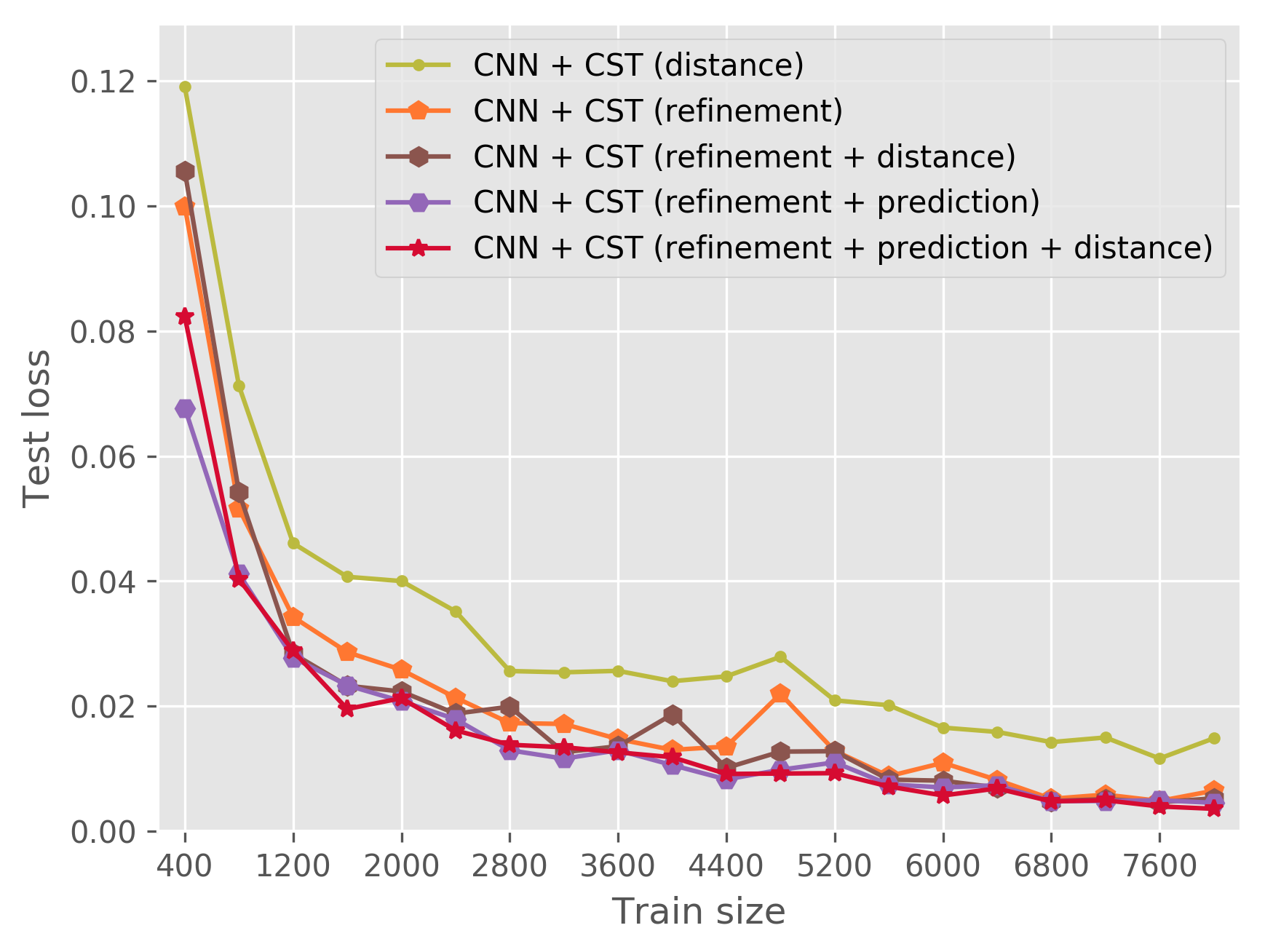}
\end{tabular}

\caption{
\label{fig:sequence_learning_curves}
Learning curves for the different setups of our architecture. The
plots show how the structural loss (Hamming distance) on the test set
changes with increasing size of the training set. On the left, our
proposed technique (red curve) is compared against its basic
components on their own, namely the CNN (blue curve) and the CST
(green curve), and against a CNN+LSTM cascade (light blue curve). On
the right, a summary of the contributions of different combinations of
features is shown.  Best viewed in color.  }

\end{figure*}

\subsection{The combined model}
After training the convolutional network, we combine it with the CST as
described in Section~\ref{sec:neural_structured}.

We train the CST model using the structured SVMs method with stochastic
subgradient descent~\cite{ratliff2007approximate} for one epoch over the
training set. The structured loss used in this task is the Hamming distance
between the predicted sequence of symbols and the true sequence.

Figure~\ref{fig:sequence_learning_curves} shows the learning curves
resulting from the experiments using the Hamming loss.  In the left
plot we compare the combined model (CNN + CST) with the single
components, i.e., the CNN and the CST in the original
variant~\cite{dragone2018pyconstruct}, and the CNN +
LSTM cascade.
The general trend is rather clear, the CNN has always an edge over the
CST, yet their combination always performs better than \emph{both}.
As expected, the gap between the CNN and the combined model is maximal
for the smallest dataset, but it remains clearly evident when the curves start
to level off (82.9\% relative error reduction at the last iteration).
The loss of the CNN + LSTM model is overall slightly higher than the one of the
CNN model, consistent with what happens for the misclassification error (see
Figure~\ref{fig:percent_errors}).

We then computed the breakdown of the CNN + CST results by features of the CST
model (see Table~\ref{tab:features_constraints}). When considering distance
only, we set its weight $\vw_{\delta}$ to a negative value, so as to minimize
the distance between the prediction of the network and the refined prediction
while satisfying the constraints, without any learning on the CST side. We then
decomposed the features of the full combined model into: [i] refinement; [ii]
refinement + distance; [iii] refinement + prediction. The right plot in
Figure~\ref{fig:sequence_learning_curves} shows the learning curves for each of
these variants
The distance feature is clearly the least effective, even if it already improves
over the CNN alone (29.6\% relative error reduction at the last iteration). The
other features perform quite similarly, but combining all features together
achieves the overall best performance.

\subsection{Comparison with DeepProbLog}

As stated in the introduction, the DeepProbLog framework~\cite{manhaeve2018deepproblog} is capable of combining neural processing with the management of logic and numeric constraints, and can in principle address a recognition task like the one presented here. Indeed, the framework proved capable of learning to add numbers represented as MNIST images~\cite{manhaeve2018deepproblog}. The equation recognition task we address here, however, is more complex, as the numbers in the equation have variable length and the position of the operators is not known a-priori. Modelling this problem with DeepProbLog requires the use of non-deterministic predicates, which are extremely expensive. Indeed, we attempted to encode the equation recognition problem in DeepProbLog\footnote{The main developer of DeepProbLog helped us in trying to optimize the encoding.}, but inference turned out to be too memory expensive to be even executed.

\section{Related work}

\paragraph{Statistical relational learning.}  Statistical relational learning (SRL)~\cite{getoor2007introduction} approaches inject prior knowledge, usually expressed in some logical formalism, into statistical learning models, with a particular emphasis on probabilistic graphical models. In SRL, the logical framework (either propositional or first order logic) is used to define the structure of the data, i.e. the known or likely relationships between the variables, which are then weighted and refined by inductive learning from data. Examples approaches include Relational Bayesian Networks~\cite{jaeger1997relational}, Markov Logic Networks~\cite{richardson2006markov}, and ProbLog~\cite{de2007problog}.  These approaches are not designed for sub-symbolic inputs.

\paragraph{Structured prediction.}  Traditional approaches to structured prediction, like conditional random fields~\cite{lafferty2001conditional} and structured output SVMs~\cite{tsochantaridis2004support} discriminatively learn an energy or scoring function over candidate input-output pairs based on a joint input-output feature map. Prior knowledge can be injected into these models through propositional (hard and soft) constraints~\cite{kristjansson2004interactive,fersini2014soft,teso2017structured}. MAP inference is solved using either ad-hoc constrained inference techniques or general solvers for combinatorial or discrete-continuous problems~\cite{roth2005integer,teso2017structured,dragone2018constructive}. These techniques, however, require one to pre-specify the relevant features (or kernel), which was shown to be suboptimal in many tasks compared to representation learning strategies.

\paragraph{Neural structured prediction.}  More recently energy-based structured prediction has been tackled using deep learning models. Notable examples include deep value networks~\cite{gygli2017deep}, structured prediction energy networks (SPENs)~\cite{belanger2016structured} and input convex neural networks~\cite{amos2017input}. These techniques implement the energy function using a neural network and use gradient methods to infer a high-quality candidate output. In contrast to their shallow alternatives, deep structured models do not straightforwardly support the addition of prior knowledge through constraints. The work by Lee~\emph{et~al.}~\cite{le2017eenforcing} addresses this issue without resorting to combinatorial search, by casting the constrained inference of a SPEN into an instance-specific learning problem with a constraint-based loss function. Unlike \method, this method is not guaranteed to 
satisfy all the constraints, especially for a complex mix of algebraic and logical ones.

\paragraph{Neuro-symbolic integration.}  Effectively combining reasoning with deep networks is still an ongoing research effort. Many attempts have been made to address this issue, most of which focus on forcing the network to learn weights that ultimately produce predictions satisfying the constraints. A popular line of research integrates constraints into the objective function using fuzzy logic, especially the \L{}ukasiewicz T-norm~\cite{diligenti2016learning,hu2016harnessing,donadello2017logic}. These approaches however cannot guarantee that the predicted outputs satisfy the hard constraints.  The same is true for the semantic loss, which encourages the output of a neural network to satisfy given constraints with high probability~\cite{xu2017semantic,di2020efficient}. Furthermore, these approaches are usually restricted to Boolean variables and logical constraints.
DeepProbLog~\cite{manhaeve2018deepproblog} extends the ProbLog  language~\cite{de2007problog} with learnable neural predicates. This powerful framework allows to reason about semantic properties of the output variables while using neural networks for low-level inputs. By building on top of a probabilistic programming framework, it inherits its ability to answer probabilistic queries other than MAP inference, something \method cannot do.  On the other hand, DeepProbLog cannot be used to model a complex structured-output prediction problem efficiently, as discussed in our experimental evaluation.

\paragraph{Declarative structured prediction.} Structured learning modulo theories~\cite{teso2017structured} is a structured learning framework dealing with (soft and hard) constraints expressed in a declarative fashion using the SMT formalism and the 
MiniZinc constraint programming language. The framework has been applied to both passive~\cite{teso2017structured,dragone2018pyconstruct} and interactive learning settings~\cite{dragone2018constructive}. Learning modulo theories shares with \method the ability to handle hybrid discrete-continuous combinatorial problems. On the other hand its underlying model is a shallow structured SVM and it thus lacks a representation learning component to deal with low-level inputs. \method can be seen as a neuro-symbolic version of this framework.

\paragraph{Other works.}  Some recent approaches have tackled problems at the intersection of constrained optimization and deep learning.  For instance, in predict-then-optimize~\cite{elmachtoub2017smart} the goal is to predict the parameters of a constrained optimization model (e.g., a scheduling problem) from data, in some cases using a deep neural network~\cite{mandi2020smart,mandi2020interior}.  In predict-then-optimize, however, the ground-truth value of the parameters to be predicted (for instance, the cost vector of a linear program) is available as supervision, whereas in structured output prediction no supervision is given on the parameters of the constraint program.  This makes the two problems conceptually very different.  Two other related threads of research are constraint learning~\cite{de2018learning} and inverse optimization~\cite{tan2019deep,tan2020learning}, which are also concerned with learning constrained optimization problems from examples.  These families of approaches, however, are not tailored for acquiring neuro-symbolic programs and it would be non-trivial to extend them to this purpose.

\section{Conclusion}

We developed a structured output prediction system combining neural
networks and constraint programming, with the aim of jointly
leveraging the effectiveness of neural networks in processing raw data
with the ability of constraint programming to deal with a wide range
of constraints over Boolean and numerical data.
The system is conceived for structured prediction tasks where examples
are scarce and prior knowledge is abundant.  We tested our approach on
one such task, namely handwritten equation recognition. A preliminary
experiment showed how CNN and even convolutional LSTM, while quickly
learning to correct syntactic errors, have serious problems in coping
with semantic errors. Our combined approach, in addition to producing
consistent predictions by design, is able to improve recognition
performance over \emph{both} the neural network and the constrained
structured model on their own, especially with smaller training sets.

We posit that our approach could be applied to several important problems in
which prior knowledge is key to their solution, such as dialogue
management~\cite{lison2015hybrid}, reinforcement learning in complex
environments~\cite{garnelo2016towards}, and interactive recommendation
systems~\cite{dragone2018constructive}, all of which we intend to pursue as
future work.

\section*{Acknowledgments}

We would like to thank Carlo Nicol\`o for running preliminary equation recognition experiments, Edoardo Battocchio for running experiments on a preliminary end-to-end pipeline, and Robin Manhaeve for helping us in trying to optimize the DeepProbLog encoding for the equation recognition problem.  The research of ST and AP was partially supported by TAILOR, a project funded by EU Horizon 2020 research and innovation programme under GA No 952215.

\bibliographystyle{kr}
\bibliography{main}
\end{document}